\documentclass[runningheads]{llncs}

\usepackage{eccv}

\usepackage{eccvabbrv}

\usepackage{graphicx}
\usepackage{booktabs}
\usepackage[accsupp]{axessibility}
\usepackage{tabularx}
\usepackage{multirow}
\usepackage{url}
\usepackage{multicol}
\usepackage{makecell}
\usepackage{diagbox}
\usepackage{wrapfig}
\usepackage{array}
\usepackage{marvosym}
\usepackage{xcolor}
\usepackage{float}

\usepackage{hyperref}

\begin{document}

\title{Divide and Conquer: Decoupled Representation Alignment for Multimodal World Models} 
\titlerunning{Decoupled Representation Alignment for Multimodal World Models}

\author{Junyuan Xiao$^{1}$ \and
Dingkang Liang$^{2,\dagger}$ \and
Xin Zhou$^{2}$ \and
Yixuan Ye$^{3}$ \and
Tongtong Su$^{4}$ \and
Guangmo Yi$^{1}$ \and
Bin Xia$^{5}$ \and
Qiang Lyu$^{6}$ \and
Shurui Shi$^{1}$ \and
Jun Huang$^{7}$ \and
Jianlou Si$^{7,\dagger}$ \and
Wenming Yang$^{1,\star}$}

\authorrunning{J.~Xiao et al.}

\institute{%
\mbox{$^{1}$Tsinghua University, China} \quad
\mbox{$^{2}$Huazhong University of Science and Technology, China}\\
\mbox{$^{3}$CSU, China} \quad
\mbox{$^{4}$ZJU, China} \quad
\mbox{$^{5}$CUHK, Hong Kong SAR, China}\\
\mbox{$^{6}$UCAS, China} \quad
\mbox{$^{7}$Alibaba Group, China}\\[4pt]
\email{xiao-jy24@mails.tsinghua.edu.cn, dkliang@hust.edu.cn}}

\maketitle
\begingroup\renewcommand\thefootnote{}%
\footnotetext{$^{\dagger}$ Project lead. \quad $^{\star}$ Corresponding author.}%
\endgroup

\begin{abstract} 
Emerging multi-modal world models attempt to jointly generate videos across diverse modalities (\eg, RGB, depth, and mask), yet they fail to fully exploit the priors of existing foundation models. We propose \textbf{M$^2$-REPA}, the first representation alignment method tailored for multi-modal video generation. Our key insight is that foundation models trained on different modality spaces naturally capture distinct domain-specific priors, acting as complementary ``experts.'' Specifically, we first decouple modality-specific features from the diffusion model's intermediate representations, then align each with its corresponding expert foundation model. To this end, we design two synergistic objectives: a multi-modal representation alignment loss that enforces feature-to-expert matching, and a modality-specific decoupling regularization that encourages complementarity across different modalities. This design enables joint optimization, exploiting priors from multiple foundation models. Extensive experiments demonstrate that our method outperforms baselines in visual quality and long-term consistency.
\end{abstract}

\section{Introduction}
\label{sec:intro}

World models enable agents to predict environmental dynamics and plan actions~\cite{ha2018world,sora}. Recent video diffusion models~\cite{denoising,hunyuanvideo,wan2025wan,cogvideox,sora,cosmos} trained on large-scale datasets with structured inputs (\eg, actions and camera movements) have emerged as promising world simulators~\cite{yang2023learning,yang2024video,qin2024worldsimbench}, with applications in autonomous driving~\cite{vista,gaia,gaia2}, embodied intelligence~\cite{llmphy,tesseract}, and interactive game engines~\cite{genie3,valevski2024diffusion,yu2025gamefactory,matrixgame20opensourcerealtime}. However, existing models operate solely on 2D RGB pixels, while the physical world is inherently 3D. This limitation poses significant challenges for 3D-aware  modeling and applications requiring accurate depth estimation and multi-modal information.

Building on this motivation, a growing body of research has explored multi-modal world models~\cite{aether,4dnex,voyager,tesseract,worldweaver}, which can simultaneously predict videos across multiple modalities such as RGB, depth, surface normals, or optical flow. These multi-modal outputs can be directly leveraged to facilitate performance enhancement mechanisms or serve various downstream tasks~\cite{worldweaver,voyager,tesseract}. Despite these promising advances, achieving robust and scalable multi-modal world models remains an open challenge.

\begin{wrapfigure}{r}{0.5\linewidth}
    \centering
    \includegraphics[width=\linewidth]{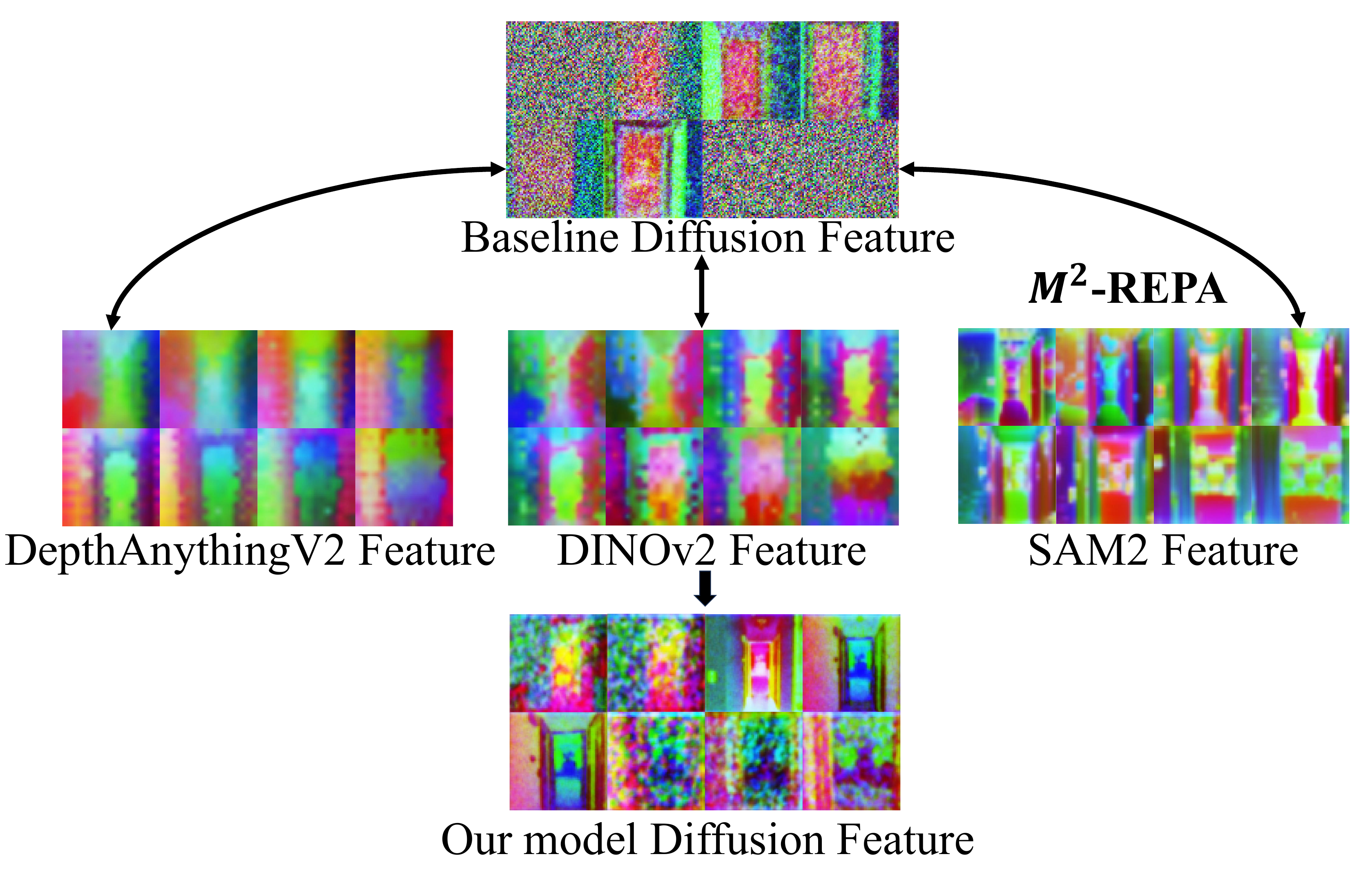} 
    \caption{\textbf{Visualization of M$^2$-REPA}. The features extracted from the backbones of DINOv2, DepthAnythingV2, and SAM2 exhibit pronounced differences, validating their distinct modality-specific characteristics. After applying our M$^2$-REPA, the extracted diffusion features align more closely with the semantic information from the foundation models.}
    \label{fig:feature}
\end{wrapfigure}

Meanwhile, recent work \cite{geometryforcing,videorepa,geometrymeetsvisionrevisiting,wristworld,visualrepresentationalignmentmultimodal} has demonstrated that leveraging high-quality representations extracted from large-scale pre-trained transformers, such as DINOv2~\cite{dinov2}, CLIP~\cite{clip}, and VGGT~\cite{vggt}, can substantially benefit video generation and downstream task performance, notably through Representation Alignment (REPA) \cite{REPA,urepa,crepa}. REPA explicitly aligns early-stage diffusion features with clean image features from self-supervised foundation vision models, providing the generation process with useful priors. This alignment allows deeper layers to focus on high-frequency content, facilitating high-fidelity generation. However, existing methods are confined to single-modality alignment using individual foundation models, leaving multi-modal video generation largely unexplored.

A key insight driving our work stems from recognizing the fundamental nature of existing foundation models: they are inherently trained to model different modality-specific spaces. For instance, DINOv2~\cite{dinov2} serves as a general-purpose feature extractor trained in the RGB visual space; DepthAnythingV2~\cite{depthanytingv2} specializes in depth estimation within the depth modality space; and SAM2~\cite{sam2} models the mask modality space for segmentation tasks. From this perspective, each foundation model can be viewed as a modality expert with distinct domain gaps and prior knowledge specific to its training modality. The differences observed between features extracted from their backbones (see Fig.~\ref{fig:feature}) provide visual evidence of this specialization, suggesting that complementary information likely exists among the prior representations from different modality-specific foundation models.

This observation motivates our work to address how foundation models can be better leveraged to improve video generation performance in the presence of multi-modal outputs. We investigate \textbf{\textit{whether simultaneously transferring complementary domain-specific priors from multiple foundation model experts across different modalities can benefit multi-modal world models.}} To validate this hypothesis, we conduct preliminary experiments on RGB-Depth (RGB-D) dual-modal video generation (Tab.~\ref{table:toy-exp}). The results reveal a clear ``modality advantage'' phenomenon: models guided by DINOv2~\cite{dinov2} alignment exhibit superior performance in the RGB modality, while those guided by DepthAnythingV2~\cite{depthanytingv2} alignment excel in depth generation. This confirms that foundation models of different modalities indeed capture unique modality-specific priors, creating opportunities for ``multi-expert collaborative guidance.'' However, naively applying multiple foundation models simultaneously to existing REPA frameworks leads to feature conflicts that hinder training optimization, yielding negligible benefits (Tab.~\ref{table:toy-exp}).
\vspace{-1mm}
\begin{table}[t]
\centering
\caption{ Quantitative results for RGB-Depth video generation on RealEstate10K (8 and 200 frames). DINOv2~\cite{dinov2} and DepthAnythingV2~\cite{depthanytingv2} excel in RGB and depth metrics, respectively. However, naively combining them for representation alignment yields marginal gains or even slight degradation compared to individual usage. \textbf{Bold} and \underline{underlined} denote the \textbf{best} and \underline{second-best} results.}

\vspace{-3mm}
\label{table:toy-exp}
{\tiny
\setlength{\tabcolsep}{2pt}
\renewcommand{\arraystretch}{1}
\begin{tabular*}{\textwidth}{@{\extracolsep{\fill}} lc|cccc|cc @{}}
\toprule
\multirow{2}{*}{\textbf{Method}} & \multirow{2}{*}{\textbf{Frames}} & \multicolumn{4}{c|}{\textbf{RGB modality}} & \multicolumn{2}{c}{\textbf{Depth modality}} \\
 & & \textbf{FVD↓} & \textbf{LPIPS↓} & \textbf{SSIM↑} & \textbf{PSNR↑} & \textbf{AbsRel↓} & \textbf{$\boldsymbol{\delta_1}$↑} \\
\midrule
RGB-D Baseline                      & 8   & 105.00 & 0.3634 & 0.5823 & 19.9183 & 0.8747 & 0.2039 \\
REPA (DINOv2~\cite{dinov2})                       & 8   & \textbf{81.38}  & \textbf{0.2239} & \underline{0.7242} & \textbf{20.1036} & \underline{0.8012} & \underline{0.4239} \\
REPA (DepthAnythingV2~\cite{depthanytingv2})              & 8   & \underline{84.12}  & \underline{0.2257} & \textbf{0.7247} & \underline{20.1023} & \textbf{0.7845} & \textbf{0.4339} \\
REPA (DINOv2+DepthAnythingV2)       & 8   & 84.56  & 0.2265 & 0.7236 & 20.0852 & 0.7768 & 0.4294 \\
\midrule
RGB-D Baseline                      & 200 & 387.50 & 0.5222 & 0.3454 & 13.7027 & 1.0155 & 0.1094 \\
REPA (DINOv2~\cite{dinov2})                       & 200 & \underline{322.25} & \underline{0.4612} & \textbf{0.4274} & \textbf{13.0289} & \underline{0.9930} & \underline{0.1330} \\
REPA (DepthAnythingV2~\cite{depthanytingv2})              & 200 & 342.25 & \textbf{0.4607} & \underline{0.4266} & \underline{12.8878} & \textbf{0.9813} & \textbf{0.1385} \\
REPA (DINOv2+DepthAnythingV2)       & 200 & \textbf{317.75} & 0.4633 & 0.4230 & 12.9625 & 0.9952 & 0.1314 \\
\bottomrule
\end{tabular*}
}
\vspace{-9mm}
\end{table}

To address these opportunities and challenges, we propose M$^2$-REPA, the first representation alignment method tailored for multi-modal video generation. M$^2$-REPA encourages models to jointly internalize domain-specific priors from diverse foundation model experts during training. To simultaneously align these priors while circumventing feature conflicts, we first decouple the diffusion latent features into distinct modality-specific representations via learnable decoupling layers, then align each representation with its corresponding expert foundation model. We design two synergistic regularization objectives: a multi-modal representation alignment loss that enforces feature-to-expert matching via cosine similarity, and a modality-specific decoupling regularization that explicitly encourages complementarity across different modalities via Centered Kernel Alignment (CKA)~\cite{CKA} similarity. This design is extensible and plug-and-play, enabling joint optimization that makes use of the priors of multi-modal experts.

Extensive experiments on challenging tri-modal (RGB-Depth-Mask) generation tasks validate our method. Whether on U-Net- or Transformer-based backbones, M$^2$-REPA consistently outperforms single-modality baselines and naive multi-expert extensions, achieving state-of-the-art fidelity in both short-term and long-range autoregressive generation.

\section{Related Work}

\subsection{Video Generation Model for World Simulation}
\noindent{\bf Interactive Long Video Generation.} Recent autoregressive video generation methods \cite{diffusionforcing,selfforcing,causvid,fifodiffusion,doft} show promise for generating arbitrarily long sequences as world models. Diffusion Forcing~\cite{diffusionforcing} employs frame-wise independent noise scheduling for autoregressive generation, while DFoT~\cite{doft} extends this with variable-length historical conditioning. The availability of large-scale video datasets with structured inputs (\eg, actions and camera poses) \cite{matrixgame20opensourcerealtime,spatialvid,omniworld,realestate10k} enables training controllable world models \cite{matrixgame20opensourcerealtime,worldmem,deepverse,genie3,valevski2024diffusion,decart2024oasis} conditioned on external commands. We introduce a novel training pipeline that integrates into existing frameworks while leveraging multi-modal generation capabilities.

\noindent{\bf Multimodal Video Generation.} Since our physical world is inherently three-dimensional, deploying world simulation and its downstream applications often necessitates information from multiple modalities. Consequently, a growing body of research has explored multi-modal world models \cite{4dnex,aether,deepverse,worldweaver,voyager,tesseract}, which generate videos beyond just RGB. For instance, TesserAct~\cite{tesseract} learns a 4D world model by training on RGB-DN (RGB, Depth, and Normal) videos. Aether~\cite{aether} develops an RGB-Depth model and leverages both modalities for reconstruction and planning. WorldWeaver~\cite{worldweaver} jointly models RGB, depth, and optical flow to enhance long video generation. While our work shares the goal of multi-modal video world modeling, it distinguishes itself by distilling strong priors from modality-specific foundation models into video representations, simultaneously improving generation quality across modalities.

\subsection{Foundation Models for Video Generation}
Foundation models such as CLIP~\cite{clip}, DINO~\cite{dinov2,dinov3}, VGGT~\cite{vggt}, DepthAnythingV2~\cite{depthanytingv2}, and SAM~\cite{sam2} learn rich, transferable representations through self-supervised learning on large-scale data, capturing robust visual semantics and geometric priors \cite{geometrymeetsvisionrevisiting,xu2025pixel,wristworld,REPA,womapworldmodelsembodied,DINO-Foresight,vjepa2,visualrepresentationalignmentmultimodal}. Recent work shows that leveraging pre-trained foundation model features substantially improves diffusion-based generation quality and consistency \cite{videorepa,REPA,crepa,xu2025pixel,geometryforcing}. The REPA family \cite{REPA,videorepa,crepa} aligns clean foundation model features with early-stage diffusion features via regularization, while Pixel-Perfect Depth~\cite{xu2025pixel} directly injects features into pixel-space diffusion blocks. However, prior works primarily focus on unimodal generation with a single foundation model. By contrast, our method exploits multiple foundation models to enhance generation quality across modalities.

\section{Preliminaries}
\subsection{Problem Formulation}

Given a single initial RGB image $\mathbf{X}^{(0)}_{\text{RGB}} \in \mathbb{R}^{3 \times H \times W}$, along with its corresponding depth map $\mathbf{X}^{(0)}_{\text{D}} \in \mathbb{R}^{1 \times H \times W}$, segmentation mask $\mathbf{X}^{(0)}_{\text{M}} \in \mathbb{R}^{C \times H \times W}$, and a sequence of structured control signals $\{\mathbf{S}^{(i)}\}_{i=1}^{T}$ (\eg, camera poses or actions), our goal is to generate temporally coherent tri-modal video sequences in an autoregressive manner. We formulate this as learning the conditional distribution:

\begin{equation}
p\left(\{\mathbf{X}^{(i)}_{\text{RGB}}, \mathbf{X}^{(i)}_{\text{D}}, \mathbf{X}^{(i)}_{\text{M}}\}_{i=1}^{T} \,\Big|\, \mathbf{X}^{(0)}_{\text{RGB}}, \mathbf{X}^{(0)}_{\text{D}}, \mathbf{X}^{(0)}_{\text{M}}, \{\mathbf{S}^{(i)}\}_{i=1}^{T}\right),
\label{eq:formulate}
\end{equation}

\noindent where $\mathbf{X}^{(i)}_{\text{RGB}}$, $\mathbf{X}^{(i)}_{\text{D}}$, and $\mathbf{X}^{(i)}_{\text{M}}$ denote the RGB frame, depth map, and segmentation mask at frame $i$, with $C$ representing the number of mask classes.

Through joint modeling of all three modalities conditioned on structured inputs, this formulation facilitates controllable generation of pixel-aligned tri-modal videos, ensuring both cross-modal spatial consistency and temporal coherence throughout the sequence.

\subsection{Autoregressive Video Diffusion Models}
Autoregressive video generation extends video sequences by predicting future frames conditioned on past observations. To model the conditional distribution defined in ~\cref{eq:formulate}, we adopt Flow Matching \cite{flowmatchinggenerativemodeling,flowstraightfastlearning}, a deterministic framework that constructs continuous trajectories from noise to target video distributions.

\noindent{\bf Flow Matching.} Let $\mathbf{x}^{(i)} = [\mathbf{X}^{(i)}_{\text{RGB}}, \mathbf{X}^{(i)}_{\text{D}}, \mathbf{X}^{(i)}_{\text{M}}] \in \mathbb{R}^{(3+1+C) \times H \times W}$ denote the concatenated tri-modal representation for the $i$-th frame. Let $\mathbf{x}_1 \in \mathbb{R}^{T \times (3+1+C) \times H \times W}$ represent the entire clean tri-modal video sequence and $\mathbf{x}_0 \sim \mathcal{N}(\mathbf{0}, \mathbf{I})$ be the standard Gaussian noise. The continuous flow timestep is denoted as $t \in [0,1]$. The linear interpolation from noise to data is defined as:

\begin{equation}
\mathbf{x}_t = t \mathbf{x}_1 + (1-t) \mathbf{x}_0,
  \label{eq:noise}
\end{equation}
with the corresponding target velocity field $\mathbf{v}_t = \mathbf{x}_1 - \mathbf{x}_0$. The model parameter $\theta$ is optimized by minimizing the flow matching objective:

\begin{equation}
  \mathcal{L}_{\text{FM}}(\theta)=\mathbb{E}_{\mathbf{x}_1, \mathbf{x}_0 \sim \mathcal{N}(\mathbf{0}, \mathbf{I}), \mathbf{y}, t \in[0,1]}\left[\left\|\mathbf{v}_\theta(\mathbf{x}_t, \mathbf{y}, t) - \mathbf{v}_t \right\|^2_2\right],
  \label{eq:flowmatching}
\end{equation}

\noindent where $\mathbf{y}$ represents all conditioning inputs, including the initial context frame $\mathbf{x}^{(0)}$ and structured signals $\{\mathbf{S}^{(i)}\}_{i=1}^{T}$. At inference, an ODE solver integrates the learned velocity field to generate samples from noise. 

\noindent{\bf Per-Frame Noise Scheduling.} To enable flexible autoregressive generation, we adapt the per-frame noise scheduling strategy from Diffusion Forcing~\cite{diffusionforcing} into the Flow Matching framework. Instead of a global timestep $t$, we assign independent continuous flow timesteps $t_i \in [0,1]$ to each frame $i$. For the video sequence $\mathbf{x}_{1:T} = \{\mathbf{x}^{(1)}, \dots, \mathbf{x}^{(T)}\}$, the intermediate state of the $i$-th frame at its specific timestep $t_i$ is constructed as:

\begin{equation}
\mathbf{x}_{t_i}^{(i)} = t_i \mathbf{x}_1^{(i)} + (1-t_i) \mathbf{x}_0^{(i)},
  \label{eq:diffusionforcing}
\end{equation}

\noindent where $\mathbf{x}_1^{(i)}$ is the clean data for frame $i$ and $\mathbf{x}_0^{(i)} \sim \mathcal{N}(\mathbf{0}, \mathbf{I})$ is its corresponding independent standard Gaussian noise. This frame-wise formulation allows the model to condition on frames at arbitrary noise levels, supporting stable generation beyond the training horizon.

\noindent{\bf Autoregressive Sampling.} At inference, we define a timestep vector $\mathbf{t} = [t_1, \dots, t_T]$. We initialize the clean context frames with $t_i = 1$ (data) for $i \leq N_{\text{ctx}}$, and the future frames to be generated with $t_j = 0$ (noise) for $j > N_{\text{ctx}}$. Sampling proceeds by solving the empirical ODE:

\begin{equation}
\mathrm{d}\mathbf{x} = \mathbf{v}_\theta\left(\mathbf{x}_{\mathbf{t}}, \mathbf{y}, \mathbf{t}\right) \mathrm{d}\mathbf{t},
\end{equation}

\noindent solved iteratively via the Euler method from $\mathbf{t} = \mathbf{0}$ to $\mathbf{t} = \mathbf{1}$ (for generated frames), extending to long-term video generation and interactive simulation.

\begin{figure*}[h]
    \centering
    \includegraphics[width=1.00\linewidth]{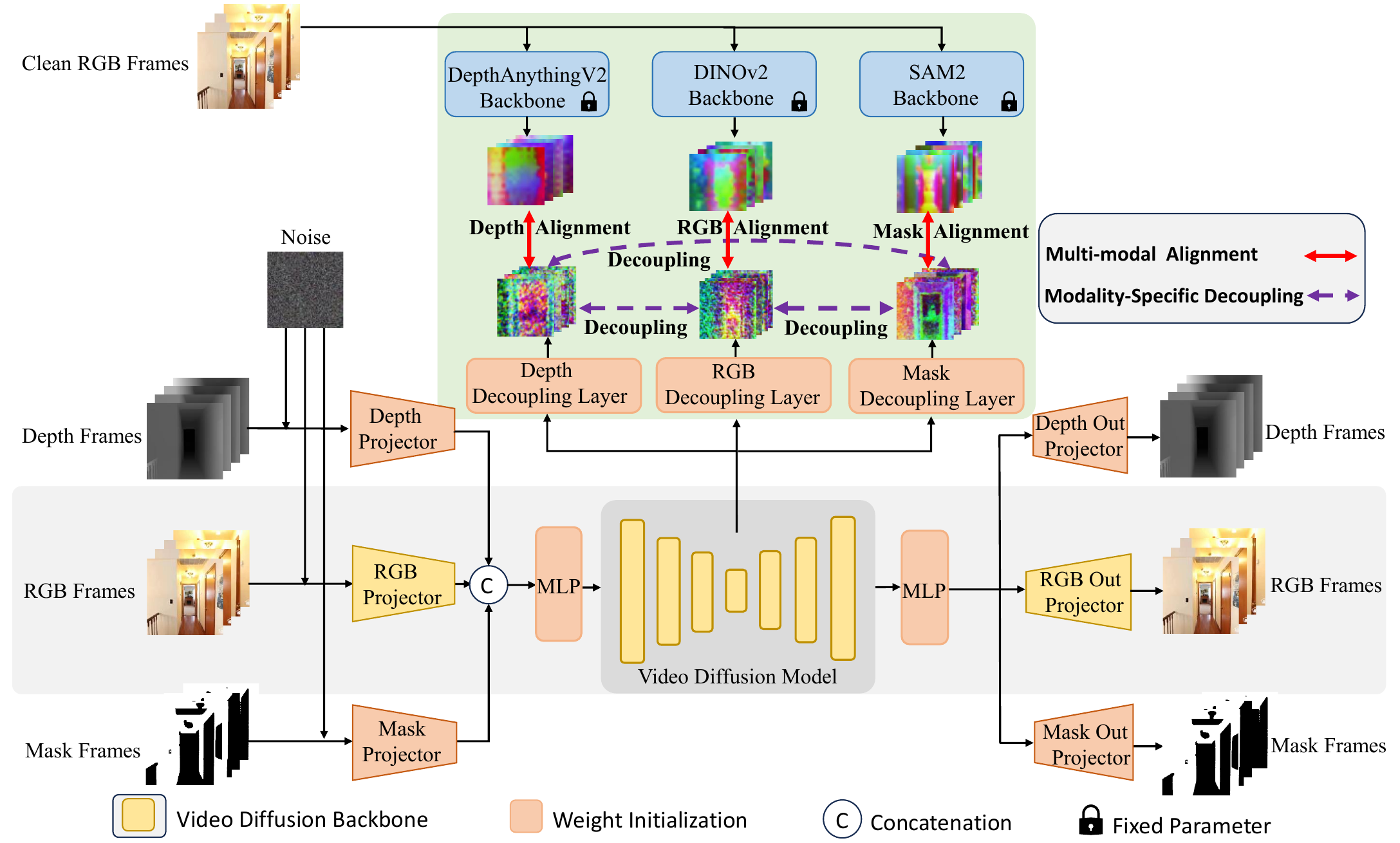}
    \caption{\textbf{Overview of M$^2$-REPA.} To mitigate feature conflicts, M$^2$-REPA decouples multi-modal features into modality-specific representations and aligns them with corresponding expert foundation models. The framework is optimized by two synergistic objectives: (1) a cosine similarity-based multi-modal alignment loss for joint representation alignment, and (2) a CKA~\cite{CKA} similarity-driven modality-specific decoupling loss to explicitly encourage cross-modal complementarity.}
    \label{fig:baselines}
\end{figure*}
\vspace{-10mm}

\section{Method}
\label{sec:method}
\subsection{Method Overview}

\noindent{\bf Motivation.} Recent works~\cite{xu2025pixel,REPA,videorepa,geometryforcing} 
have explored leveraging features from pre-trained foundation encoders, such as DINOv2~\cite{dinov2}, 
CLIP~\cite{clip}, and VGGT~\cite{vggt}, to ground the generation process and improve temporal consistency in video
diffusion models. However, these methods are still largely confined to RGB-only generation and typically rely on isolated foundation models, leaving multi-modal scenarios underexplored. We observe that existing
foundation models are trained in distinct modality spaces: DINOv2~\cite{dinov2} captures rich RGB visual representations, DepthAnythingV2~\cite{depthanytingv2} excels at geometric depth reasoning, and SAM2~\cite{sam2} provides 
strong segmentation knowledge. These models therefore serve as modality experts with unique and complementary priors. The distributional differences among their backbone features
(see~\cref{fig:feature}) further reveal their specialization. This insight motivates our key question: can a multi-modal world model benefit from simultaneously transferring domain-specific priors from multiple foundation model experts across different modalities?

\noindent{\bf Observation.} As a pilot study, we first construct a dual-modality video diffusion 
model for joint RGB-depth generation by concatenating both modalities, applying MLP projection 
layers, and fine-tuning with modality-specific heads atop a pre-trained backbone~\cite{doft}. 
We conduct experiments using the REPA alignment method with two foundation models: DINOv2 (RGB expert) and Depth 
Anything V2 (depth expert). As shown in~\cref{table:toy-exp}, a clear \textbf{``modality advantage''} 
emerges: the model guided by the RGB expert (DINOv2) achieves superior generation quality in the RGB modality, while the model guided by the depth expert (DepthAnythingV2) exhibits stronger performance in the depth modality. This demonstrates that foundation models trained on different modalities capture distinct modality-specific prior knowledge. \textbf{This finding reveals a key opportunity}: by simultaneously transferring knowledge from multiple expert foundation models into the video diffusion model, we can improve high-fidelity generation across multiple modalities, thereby achieving multi-expert guided generation.

Therefore, we propose a representation alignment-based approach that first disentangles modality-specific features within the video diffusion model, then aligns each with its corresponding expert foundation model. This encourages the network to jointly internalize diverse prior knowledge during training. In the following sections, we detail our M$^2$-REPA framework, which bridges the gap between these decoupled diffusion features and heterogeneous foundation priors via two carefully designed regularization objectives.

\subsection{MultiModal Representation Alignment}
\noindent{\bf Challenge.}\phantomsection\label{para:challenge} 
Introducing priors from multiple modality-specific foundation models 
into Video Diffusion Models (VDMs) is non-trivial. Existing strategies fall into two 
categories: direct injection~\cite{xu2025pixel} and implicit alignment~\cite{crepa,REPA,
urepa,videorepa,geometryforcing}. Direct injection suffers from the absence of video 
inputs at inference and inherent incompatibility between pixel-space foundation features 
and VAE latent spaces; injecting features from multiple modalities further risks feature 
redundancy and training collapse. REPA~\cite{REPA} offers a compelling alternative by 
regularizing alignment between diffusion and foundation model features during training, 
improving generation quality without inference-time overhead. However, existing REPA 
methods~\cite{REPA,urepa,crepa,videorepa} are confined to single-modality RGB generation 
with a single foundation model. Moreover, naively extending alignment to multiple foundation 
models via feature similarity alone introduces feature conflicts, yielding negligible 
gains (see~\cref{table:toy-exp}).

These limitations collectively indicate that directly applying standard REPA is inadequate for enhancing multi-modal world model generation. This requires a different strategy that simultaneously
incorporates priors from multiple modality-specific foundation models while actively 
preventing feature conflicts arising from concurrent multi-modal alignment.

To address these challenges, we propose \textbf{M$^2$-REPA} (see~\cref{fig:baselines}), 
which extracts multi-modal features from intermediate diffusion representations, decouples 
them through modality-specific decoupling layers, and aligns each with its corresponding ``single-modality expert'' foundation model. We introduce two complementary regularization objectives: (1)~\emph{Multi-modal
Representation Alignment} and (2)~\emph{Modality-specific Feature Decoupling}, which
jointly optimize the alignment between diffusion latent features and multiple pre-trained
foundation models, making use of their complementary modality-specific priors.

\noindent{\bf Model Architecture.}\phantomsection\label{sec:model_arch}
To accommodate diverse generation scenarios, we construct a unified tri-modal (RGB-Depth-Mask) video diffusion baseline. Without loss of generality, we detail the pixel-space instantiation built upon the autoregressive UViT backbone~\cite{uvit2} below, while deferring the structural adaptations for the latent-space DiT variant to the supplementary material. As illustrated in \cref{fig:baselines}, we introduce separate projection layers for each modality. Specifically, we duplicate the pre-trained RGB input and output projection layers to serve as dedicated encoders and decoders for the newly incorporated Depth and Mask modalities. For each modality $m \in \{\text{RGB}, \text{D}, \text{M}\}$, we first encode the input data $\mathit{X}_t^m$ as $\mathit{e}_t^m = \text{Enc}^m(\mathit{X}_t^m)$. The tri-modal embeddings $\mathit{e}_t^m \in \mathbb{R}^{d}$ are then concatenated along the channel dimension and projected through a learnable MLP layer to obtain a unified joint representation $\mathit{e}_t = \text{MLP}_{\text{fuse}}([\mathit{e}_t^{\text{RGB}}; \mathit{e}_t^{\text{D}}; \mathit{e}_t^{\text{M}}])$. The fused embeddings are subsequently processed by the diffusion backbone $f_\theta$, conditioned on context frames and task-specific signals $\mathit{y}$ (\eg, camera poses), to predict the velocity field $\tilde{\mathit{v}}_t = f_\theta(\mathit{e}_t, \mathit{y}, t)$. The output is projected via another MLP and split channel-wise to recover modality-specific representations $[\hat{\mathit{e}}_t^{\text{RGB}}; \hat{\mathit{e}}_t^{\text{D}}; \hat{\mathit{e}}_t^{\text{M}}] = \text{Split}(\text{MLP}_{\text{split}}(\tilde{\mathit{v}}_t))$. Finally, each token is decoded to produce the final tri-modal velocity field $\hat{\mathit{v}}_t = [\hat{\mathit{v}}_t^{\text{RGB}}, \hat{\mathit{v}}_t^{\text{D}}, \hat{\mathit{v}}_t^{\text{M}}]$, allowing us to minimize the objective in~\cref{eq:flowmatching}.

\noindent{\bf Multi-modal Representation Alignment.}
While prior REPA-based methods~\cite{videorepa,crepa} focus on RGB modality alignment using a single vision foundation model, we introduce Multi-modal Representation Alignment (M$^2$-REPA), the first alignment approach for multi-modal video generation that simultaneously leverages multiple foundation models. By exploiting complementary strengths of diverse pre-trained models, our method improves generation quality and semantic consistency.

We enforce alignment between the diffusion intermediate features $h_t = f_\theta(x_t)$ extracted from the denoising network on noisy frames $x_t$ and the target representations $y^{(k)}_* = f^{(k)}(x)$ extracted from $K$ pre-trained foundation models on clean frames $x$, where $k \in \{1, ..., K\}$ denotes the model index and $K$ is the total number of foundation models. In this work, we employ three complementary ``modality experts'' ($K=3$): DINOv2~\cite{dinov2} for RGB, DepthAnythingV2~\cite{depthanytingv2} for depth, and SAM2~\cite{sam2} for mask segmentation.

For each modality $k$, features are reshaped to $\mathbb{R}^{BT \times N \times D}$ where $B$ is batch size, $T$ is frame count, and $N$ is spatial tokens. After L2 normalization, the multi-modal alignment loss is defined as:

\begin{equation}
\mathcal{L}_{\text{M$^2$-REPA}} = -\frac{1}{K}\sum_{k=1}^{K} \mathbb{E}\left[\frac{1}{N}\sum_{n=1}^{N} \text{cos}(\hat{y}^{(k)}_{*,[n]}, \hat{h}^{(k)}_{\phi,[n]})\right],
\label{eq:mm-repa}
\end{equation}

\noindent where $\hat{h}^{(k)}_\phi = g^{(k)}_\phi(h_t)$ denotes the projected features obtained through a lightweight MLP $g^{(k)}_\phi$(detailed in the next subsection), $\text{cos}(\cdot, \cdot)$ represents the cosine similarity, and the loss maximizes token-wise cosine similarity across all modalities. During training, only the MLP projectors and the diffusion backbone are optimized while all foundation models remain frozen.

\noindent{\bf Modality-specific Feature Decoupling.}
However, naively aligning diffusion intermediate features with multiple foundation models simultaneously can paradoxically degrade generation performance compared to using a single foundation model (see \cref{table:toy-exp}). This stems from the inherent conflicts between different modality-specific priors: directly forcing the shared diffusion features to match heterogeneous foundation model representations leads to feature entanglement and optimization difficulties. To address this challenge, we introduce Modality-specific Feature Decoupling, which explicitly disentangles modality-specific information from the diffusion backbone before alignment, enabling effective multi-modal joint optimization.

Concretely, we introduce modality-specific decoupling layers $\{g^{(k)}_\phi\}_{k=1}^K$ implemented as lightweight MLPs, which project the shared diffusion features $h_t$ into modality-specific subspaces: $\hat{h}^{(k)}_\phi = g^{(k)}_\phi(h_t)$. Each decoupled representation $\hat{h}^{(k)}_\phi$ is then aligned with its corresponding foundation model target $y^{(k)}_*$ via Eq.~\ref{eq:mm-repa}. This design is simple yet highly effective, allowing each modality expert to guide its dedicated feature subspace without interfering with others. Empirical analysis and further discussions motivating the choice of MLPs for feature decoupling are deferred to the supplementary material.

To further encourage the decoupled features to capture complementary information across modalities, we design a Modality-specific Decoupling Regularization Loss. We employ Centered Kernel Alignment (CKA)~\cite{CKA} as the similarity metric, which is particularly well-suited for quantifying representational similarity due to its invariance to isotropic scaling and orthogonal transformations. Specifically, we minimize the pairwise CKA similarity among all $K$ decoupled modality-specific features to encourage orthogonality and complementarity. The modality-specific decoupling regularization loss is defined as:

\begin{equation}
\mathcal{L}_{\text{decouple}} = \frac{2}{K(K-1)}\sum_{i=1}^{K}\sum_{j=i+1}^{K} \text{CKA}(\hat{h}^{(i)}_\phi, \hat{h}^{(j)}_\phi),
\label{eq:decouple}
\end{equation}

\noindent where $\hat{h}^{(k)}_\phi$ denotes the decoupled feature for the $k$-th modality (e.g., RGB, Depth, and Mask when $K=3$). By minimizing $\mathcal{L}_{\text{decouple}}$, we explicitly penalize redundant information sharing across modalities, ensuring that each modality-specific subspace captures distinct and complementary semantic structures.

\subsection{MultiModal Video Diffusion Models}
The complete training objective integrates flow matching, multi-modal alignment, and decoupling regularization:

\begin{equation}
\mathcal{L}_{\text{total}} = \mathcal{L}_{\text{FM}} + \lambda_{\text{align}} \mathcal{L}_{\text{M$^2$-REPA}} + \lambda_{\text{decouple}} \mathcal{L}_{\text{decouple}},
\label{eq:total_loss}
\end{equation}

\noindent where $\lambda_{\text{align}}$ and $\lambda_{\text{decouple}}$ balance the contributions of alignment and decoupling objectives. During training, only the diffusion backbone and MLP projectors $\{g^{(k)}_\phi\}_{k=1}^K$ are trainable, while all foundation models remain frozen to preserve their pre-trained knowledge. This design enables our M$^2$-REPA framework to harness diverse foundation model priors without feature conflicts, improving multi-modal video generation quality. Hyperparameter sensitivity analysis is deferred to the supplementary material.

\section{Experiments}

\begin{table*}[t]
\centering
\caption{
Quantitative comparison on the RealEstate10K dataset for both short-term (8-frame) and long-term (200-frame) video generation. Our M$^2$-REPA achieves the best performance across most metrics.}
\label{table:Quantitative comparison}
{\tiny
\setlength{\tabcolsep}{2pt}
\renewcommand{\arraystretch}{1}
\begin{tabular*}{\textwidth}{@{\extracolsep{\fill}} lc|cccc|cc|c @{}}
\toprule
\multirow{2}{*}{\textbf{Method}} & \multirow{2}{*}{\textbf{Frames}} & \multicolumn{4}{c|}{\textbf{RGB modality}} & \multicolumn{2}{c|}{\textbf{Depth modality}} & \textbf{Mask modality} \\
 & & \textbf{FVD↓} & \textbf{LPIPS↓} & \textbf{SSIM↑} & \textbf{PSNR↑} & \textbf{AbsRel↓} & \textbf{$\boldsymbol{\delta_1}$↑} & \textbf{mIoU↑} \\
\midrule
DFoT~\cite{doft}                            & 8   & 110.88 & 0.2471 & 0.6691 & 18.3858 & -- & -- & -- \\
Geometry Forcing~\cite{geometryforcing}                & 8   & 106.00 & \underline{0.2292} & 0.7008 & 18.8756 & -- & -- & -- \\
RGB-DM Baseline                 & 8   & 102.44 & 0.3449 & 0.5906 & 19.5264 & 0.9076 & 0.1724 & \underline{0.8839} \\
REPA (DINOv2~\cite{dinov2})                   & 8   & 100.00 & 0.2372 & 0.7142 & \underline{19.6205} & 1.0329 & 0.2987 & 0.8467 \\
REPA (SAM2~\cite{sam2})                     & 8   & 97.63  & 0.2344 & \underline{0.7172} & 19.5104 & 0.8431 & 0.2866 & 0.8801 \\
REPA (DepthAnythingV2~\cite{depthanytingv2})          & 8   & \underline{96.81}  & 0.2364 & 0.7165 & 19.5978 & \underline{0.8301} & \underline{0.3090} & 0.8787 \\
M$^2$-REPA (ours)         & 8   & \textbf{81.13}  & \textbf{0.2123} & \textbf{0.7439} & \textbf{20.4845} & \textbf{0.8224} & \textbf{0.4449} & \textbf{0.9045} \\
\midrule
DFoT~\cite{doft}                            & 200 & 401.50 & 0.5218 & 0.3536 & 11.1614 & -- & -- & -- \\
Geometry Forcing~\cite{geometryforcing}                & 200 & 371.50 & 0.4897 & 0.3791 & 11.2901 & -- & -- & -- \\
RGB-DM Baseline                 & 200 & 407.50 & 0.5334 & 0.3391 & 12.7900 & 0.9991 & 0.1238 & 0.6385 \\
REPA (DINOv2~\cite{dinov2})                   & 200 & 308.50 & 0.4682 & \textbf{0.4417} & \textbf{12.9255} & 1.0007 & 0.1240 & 0.6191 \\
REPA (SAM2~\cite{sam2})                     & 200 & \underline{301.25} & 0.4696 & 0.4319 & 12.5578 & \underline{0.9958} & \underline{0.1247} & 0.6283 \\
REPA (DepthAnythingV2~\cite{depthanytingv2})          & 200 & 319.00 & \underline{0.4668} & \underline{0.4390} & 12.8776 & 0.9988 & 0.1240 & \underline{0.6446} \\
M$^2$-REPA (ours)         & 200 & \textbf{262.50} & \textbf{0.4592} & 0.4374 & \underline{12.8918} & \textbf{0.9813} & \textbf{0.1408} & \textbf{0.6591} \\
\bottomrule
\end{tabular*}
}
\end{table*}

\begin{table*}[t]
\centering
\caption{
Ablation study of M$^2$-REPA components. $^\dagger$D+S+DA denotes the naive combination of DINOv2, SAM2, and DepthAnythingV2.
}
\label{table:ablation}
{\tiny
\setlength{\tabcolsep}{2pt} 
\begin{tabular}{lc|cccc|cc|c}
\toprule
\multirow{2}{*}{\textbf{Method}} & \multirow{2}{*}{\textbf{Frames}} & \multicolumn{4}{c|}{\textbf{RGB modality}} & \multicolumn{2}{c|}{\textbf{Depth modality}} & \textbf{Mask modality} \\
 & & \textbf{FVD↓} & \textbf{LPIPS↓} & \textbf{SSIM↑} & \textbf{PSNR↑} & \textbf{AbsRel↓} & \textbf{$\boldsymbol{\delta_1}$↑} & \textbf{mIoU↑} \\
\midrule
RGB-DM Baseline                              & 8   & 102.44 & 0.3449 & 0.5906 & 19.5264 & 0.9076 & 0.1724 & \underline{0.8839} \\
REPA (DINOv2~\cite{dinov2})                  & 8   & 100.00 & 0.2372 & 0.7142 & \underline{19.6205} & 1.0329 & 0.2987 & 0.8467 \\
REPA (SAM2~\cite{sam2})                      & 8   & 97.63  & \underline{0.2344} & \underline{0.7172} & 19.5104 & 0.8431 & 0.2866 & 0.8801 \\
REPA (DepthAnythingV2~\cite{depthanytingv2}) & 8   & 96.81  & 0.2364 & 0.7165 & 19.5978 & \underline{0.8301} & \underline{0.3090} & 0.8787 \\
REPA (D+S+DA)$^\dagger$           & 8   & \underline{96.38}  & 0.2377 & 0.7133 & 19.4952 & 0.8384 & 0.3095 & 0.8648 \\
M$^2$-REPA (cos$^2$ loss)              & 8   & 99.56  & 0.2381 & 0.7147 & 19.4227 & 0.8822 & 0.3103 & 0.8637 \\
M$^2$-REPA (CKA loss, Ours)            & 8   & \textbf{81.13}  & \textbf{0.2123} & \textbf{0.7439} & \textbf{20.4845} & \textbf{0.8224} & \textbf{0.4449} & \textbf{0.9045} \\
\midrule
RGB-DM Baseline                              & 200 & 407.50 & 0.5334 & 0.3391 & 12.7900 & 0.9991 & 0.1238 & 0.6385 \\
REPA (DINOv2~\cite{dinov2})                  & 200 & 308.50 & 0.4682 & \underline{0.4417} & \underline{12.9255} & 1.0007 & 0.1240 & 0.6191 \\
REPA (SAM2~\cite{sam2})                      & 200 & \underline{301.25} & 0.4696 & 0.4319 & 12.5578 & 0.9958 & 0.1247 & 0.6283 \\
REPA (DepthAnythingV2~\cite{depthanytingv2}) & 200 & 319.00 & 0.4668 & 0.4390 & 12.8776 & 0.9988 & 0.1240 & 0.6446 \\
REPA (D+S+DA)$^\dagger$           & 200 & 328.75 & 0.4689 & 0.4340 & 12.7331 & \underline{0.9940} & \underline{0.1275} & 0.6470 \\
M$^2$-REPA (cos$^2$ loss)              & 200 & 319.75 & \underline{0.4652} & \textbf{0.4420} & \textbf{12.9756} & 0.9992 & 0.1239 & \underline{0.6490} \\
M$^2$-REPA (CKA loss, Ours)            & 200 & \textbf{262.50} & \textbf{0.4592} & 0.4374 & 12.8918 & \textbf{0.9813} & \textbf{0.1408} & \textbf{0.6591} \\
\bottomrule
\end{tabular}
}
\end{table*}

\noindent{\bf Experimental Setup.} To evaluate M$^2$-REPA, we build our 
multi-modal baseline upon DFoT~\cite{doft}, following its experimental protocol. 
We consider two representative scenarios: \textbf{(1) Real-world scene generation} 
on RealEstate10K~\cite{realestate10k} under camera-pose conditioning, and 
\textbf{(2) Dynamic environment generation} in Minecraft~\cite{minecraft} under action conditioning. 
For multi-modal data construction, depth pseudo-labels are estimated via DepthAnythingV2~\cite{depthanytingv2}, and segmentation masks are extracted using 
SAM2~\cite{sam2}, retaining the top-3 highest-confidence masks per frame. 
We adopt UViT~\cite{uvit2} as the diffusion backbone for RealEstate10K
and DiT~\cite{dit} for Minecraft, spanning both U-Net-based and Transformer-based
architectures to evaluate our approach
across diverse backbone designs.

\begin{figure}[t]
    \centering
    \includegraphics[width=0.8\linewidth]{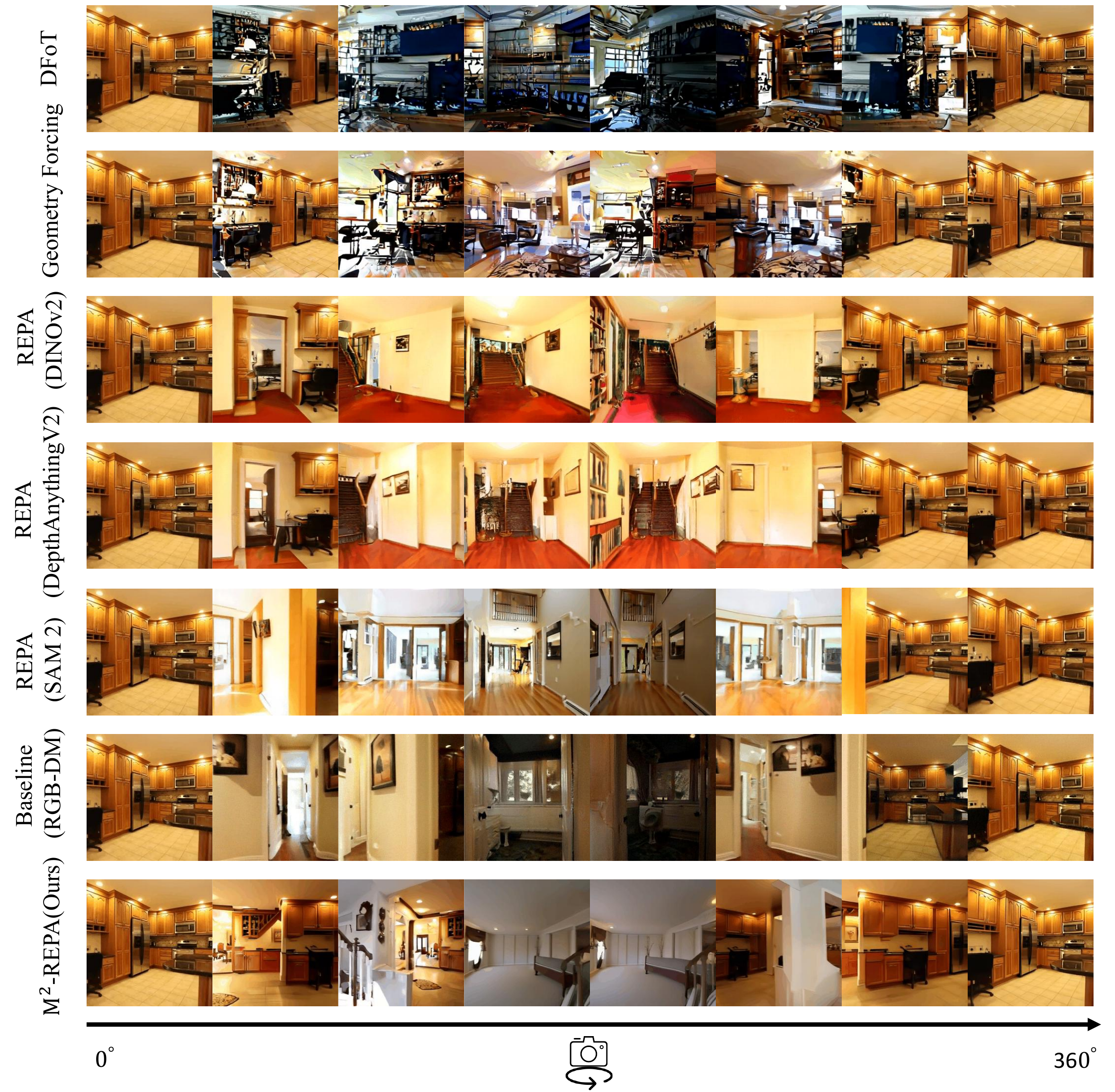}
    \caption{Qualitative comparison of camera view-conditioned video generation under fullcircle rotation. Videos are generated from a input frame and corresponding per-frame camera poses simulating a full 360° rotation.}
    \label{fig:rotate}
\end{figure}
\noindent{\bf Implementation Details.} Guided by the architecture in \cref{sec:model_arch}, we evaluate our method on the challenging joint RGB-Depth-Mask generation task. For camera view-conditioned video generation~\cite{realestate10k}, the model is trained in pixel space using 8-frame video clips at $256 \times 256$ resolution for 2,500 iterations. For action-conditioned video generation~\cite{minecraft}, we adopt a latent-diffusion paradigm where videos are pre-encoded into a latent space via a pre-trained VAE~\cite{vae}; training is conducted on 50-frame latent sequences for 5,000 iterations. Both setups share a learning rate of $8 \times 10^{-6}$ and a batch size of 8. During inference, we perform auto-regressive roll-outs conditioned on the first frame, encompassing both short-range (8 frames) and long-range (200 frames) horizons. To balance the multi-objective optimization, we empirically set $\lambda_{\text{align}}=0.5$ and $\lambda_{\text{decouple}}=0.05$. All experiments are executed on a cluster of 8 NVIDIA H20 GPUs.

\noindent{\bf Evaluation Metrics.}
For the RGB modality, we adopt standard video generation metrics to assess visual quality, including FVD~\cite{fvd}, PSNR, SSIM~\cite{psnr} and LPIPS~\cite{lpips}. For the depth modality, following DepthCrafter~\cite{depthcrafter}, we first align predicted depth maps with ground truth via uniform scale-shift normalization across the entire video, then evaluate geometric accuracy using AbsRel and $\delta_1$ accuracy~\cite{depthcrafter,depthanytingv2}. For segmentation masks, we compute mIoU~\cite{miou} by performing greedy one-to-one matching between predicted and ground-truth masks, averaging matched IoU scores at the frame level, then across all frames. Additional metric details are provided in the supplementary material.

\subsection{Quantitative comparisons}

\noindent{\bf Real-world Scene Generation.} We conduct comprehensive quantitative evaluations on the RealEstate10K dataset~\cite{realestate10k}, covering both short-term (8-frame) and long-term (200-frame) video generation tasks. As shown in Tab.~\ref{table:Quantitative comparison}, we compare M$^2$-REPA against several strong baselines, including DFoT~\cite{doft}, Geometry Forcing~\cite{geometryforcing}, our RGB-DM baseline, and single-modality REPA variants that leverage DINOv2~\cite{dinov2}, SAM2~\cite{sam2}, and DepthAnythingV2~\cite{depthanytingv2}. From the metrics, our method achieves substantial improvements across all modalities. M$^2$-REPA consistently outperforms all baselines across multiple evaluation metrics spanning three modalities, including FVD, $\delta_1$, and mIoU, demonstrating the best generation performance. These results validate that our approach harnesses the complementary priors from multiple foundation models, enabling improved joint generation and optimization across modalities.

\noindent{\bf Dynamic Environment Generation.} To evaluate the generalizability of our framework across diverse data distributions and architectures, we extend our tri-modal baseline to the dynamic and visually varied Minecraft environment~\cite{minecraft}. For this task, we employ DiT~\cite{dit} as the underlying backbone to further validate our method's backbone-agnostic nature. As summarized in \cref{tab:repa_variants}, M$^2$-REPA yields notable performance gains in long-term (150-frame) video generation. These results indicate that our method can be integrated into various video diffusion frameworks, delivering consistent gains regardless of the specific data domain or network architecture.

\begin{table}[t]
\centering
\caption{\textbf{Evaluation on dynamic environment generation in Minecraft.} $^\dagger$D+S+DA denotes the naive combination of DINOv2, SAM2, and DepthAnythingV2.}
{\tiny
\setlength{\tabcolsep}{2pt}
\renewcommand{\arraystretch}{1}
\begin{tabular*}{\textwidth}{@{\extracolsep{\fill}} lccccccccc @{}}
\toprule
\textbf{Method} & \textbf{Frames} & \textbf{FVD↓} & \textbf{LPIPS↓} & \textbf{SSIM↑} & \textbf{PSNR↑} & \textbf{AbsRel↓} & \textbf{$\boldsymbol{\delta_1}$↑} & \textbf{mIoU↑} \\
\midrule
RGB-DM Baseline & 150 & 191.25 & 0.5561 & \underline{0.4515} & 12.8364 & \underline{1.0467} & 0.0228 & 0.5962 \\
REPA(DINOv2) & 150 & \underline{129.50} & \underline{0.5300} & 0.4482 & \underline{12.9883} & 1.0505 & 0.0228 & 0.6371 \\
REPA(SAM2) & 150 & 132.12 & 0.5386 & 0.4474 & 12.9483 & 1.0539 & \textbf{0.0229} & \underline{0.6455} \\
REPA(DepthAnythingV2) & 150 & 133.50 & 0.5390 & 0.4474 & 12.9510 & 1.0545 & \textbf{0.0229} & 0.6451 \\
REPA (D+S+DA)$^\dagger$ & 150 & 144.75 & 0.5388 & 0.4459 & 12.9494 & 1.0562 & \textbf{0.0229} & 0.6369 \\
M$^2$-REPA (Ours) & 150 & \textbf{81.81} & \textbf{0.5258} & \textbf{0.4566} & \textbf{13.0081} & \textbf{1.0159} & \underline{0.0228} & \textbf{0.6612} \\
\bottomrule
\end{tabular*}
}
\label{tab:repa_variants}
\end{table}

\subsection{Qualitative comparisons}

\begin{figure}[t]
    \centering
    \includegraphics[width=0.86\linewidth]{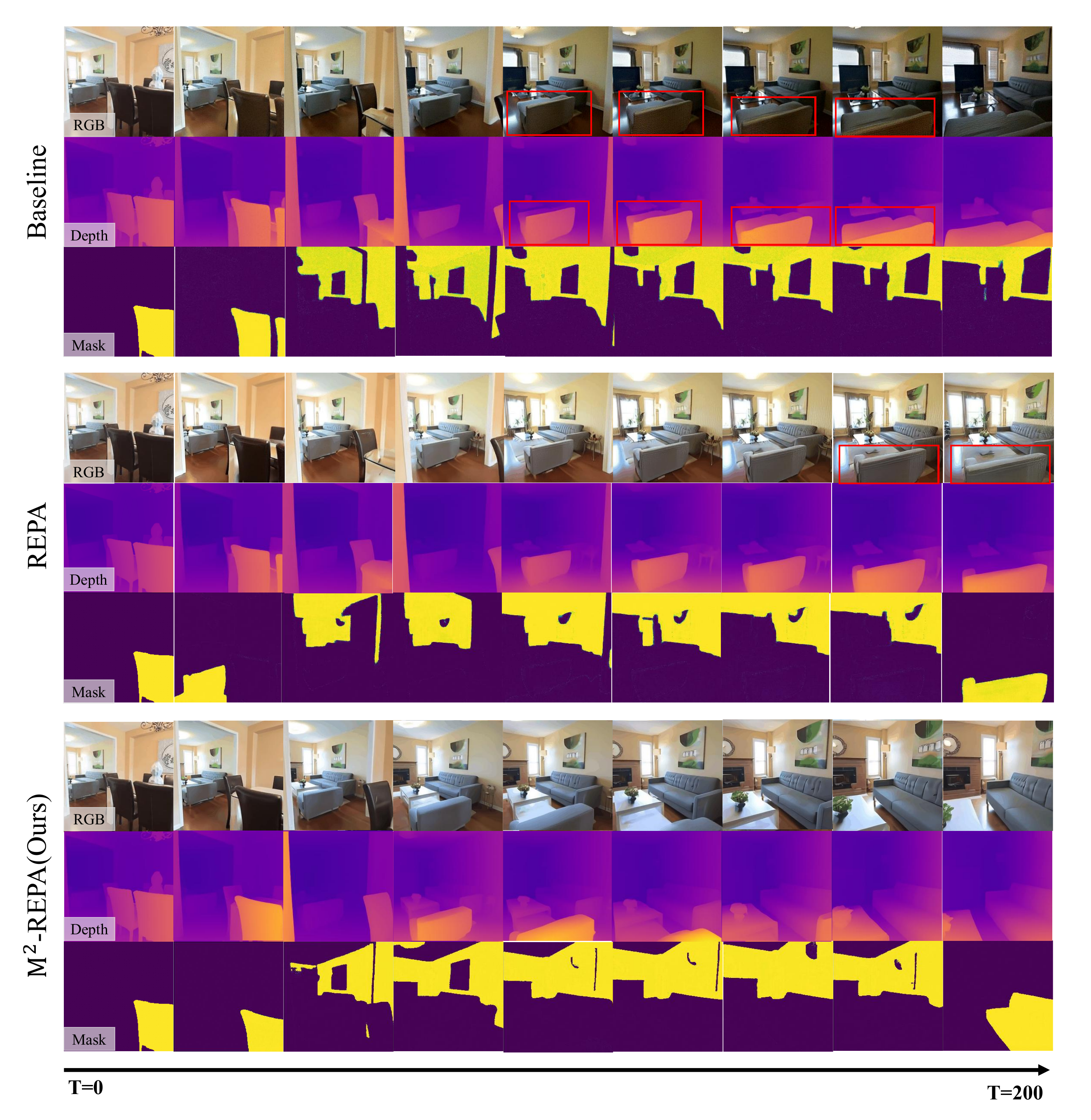}
    \caption{Qualitative comparison in ablation study. Under the 200-frame long video setting, we compare our method (M$^2$-REPA) against baseline and REPA methods.}
    \label{fig:rgbdseg-visual}
\end{figure}

Fig.~\ref{fig:rotate} presents qualitative comparisons of our 
M$^2$-REPA against baseline methods on the RealEstate10K dataset. Each video is generated conditioned on the first frame and future camera poses, simulating a 360$^{\circ}$ rotation. Examining the four intermediate frames, we observe that both DFoT and Geometry Forcing suffer from severe scene collapse, while the RGB-DM baseline and single-modality REPA variants exhibit varying degrees of distortion and blurry artifacts. In contrast, our method generates RGB frames with sharper details and more realistic textures. Additionally, Fig.~\ref{fig:rgbdseg-visual} visualizes the generation results across all three modalities (RGB-Depth-Mask) under the long video generation setting (200 frames). While the baseline and single-modality REPA methods exhibit memory drift of scene elements such as the sofa during long-horizon generation, M$^2$-REPA maintains consistent temporal coherence throughout the sequence, demonstrating substantial joint quality improvements across all modalities.

\vspace{-3mm}

\subsection{Ablation studies}

\noindent{\bf Effect of Naive Multi-Modal Alignment.} As shown in Tab.~\ref{table:ablation}, while individual single-modality REPA variants consistently outperform the RGB-DM baseline, naively aligning all three foundation models simultaneously without modality-specific decoupling severely degrades long-video generation. This phenomenon corroborates our core hypothesis (see \cref{para:challenge}): directly forcing shared representations to match heterogeneous foundation models inevitably triggers severe feature conflicts.

\noindent{\bf Effectiveness of Modality-Specific Decoupling Regularization.}
The results demonstrate that incorporating modality-specific decoupling regularization yields substantial performance improvements compared to the naive direct alignment of multiple foundation models. To further assess the importance of our modality-specific decoupling loss $\mathcal{L}_{\text{decouple}}$, we compare two regularization strategies: (1) \textbf{cos$^2$ loss}, which minimizes $\frac{1}{3}\sum_{i,j}(\text{cos}(\hat{h}^{(i)}_\phi, \hat{h}^{(j)}_\phi))^2$ to encourage orthogonality, and (2) \textbf{CKA loss}~\cite{CKA} (our method), as defined in Equation~\ref{eq:decouple}. The empirical results reveal a clear advantage of the CKA-based formulation. These findings underscore the importance of employing CKA as a similarity metric, which exhibits superior robustness to feature distribution variations and orthogonal transformations compared to naive cosine-based penalties.

\begin{figure}[t]
  \centering
  \includegraphics[width=0.8\linewidth]{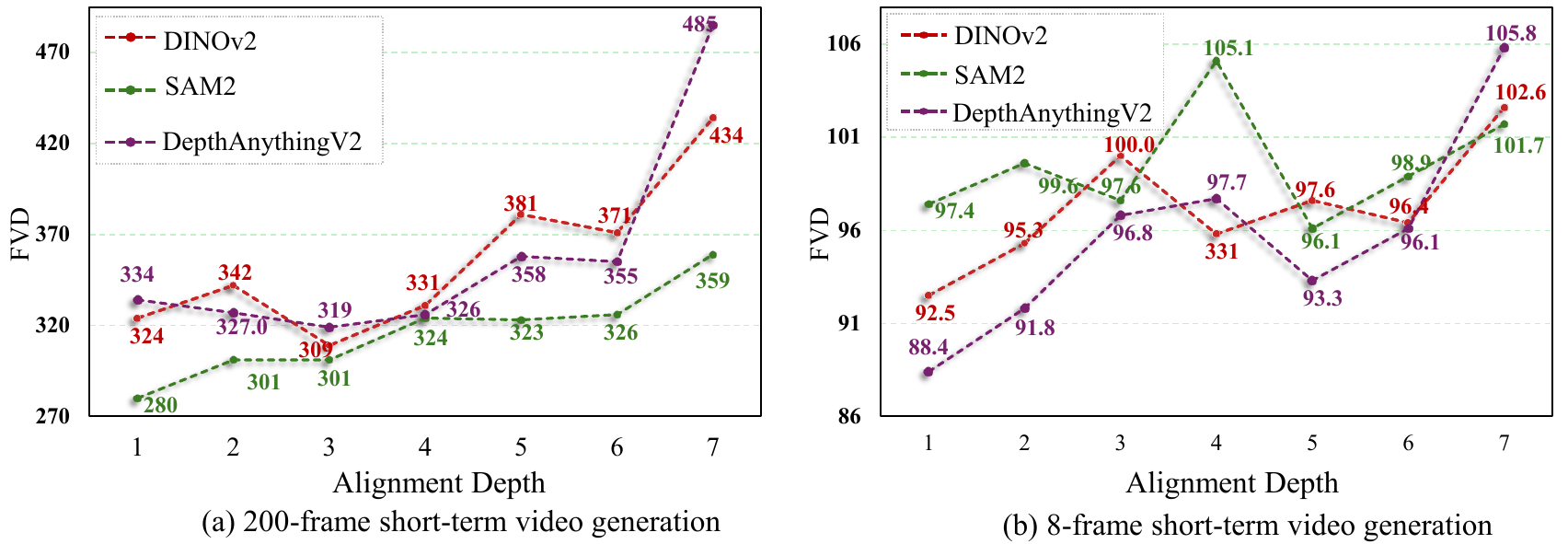}
\caption{\textbf{Ablation on alignment layer depth.} FVD-200 and FVD-8 scores 
across different alignment layers of the diffusion backbone.}
   \label{fig:layer_selection}
\end{figure}

\noindent{\bf Feature Layer Selection and Sensitivity.}
 For foundation models, we strictly follow the original settings from REPA series works~\cite{REPA, urepa}: DINOv2 uses the 4th layer output, DepthAnythingV2 uses the last layer of the encoder, and SAM2 uses the 11th layer of the image encoder. Additionally, as shown in Fig.~\ref{fig:layer_selection}, we ablate the choice 
of alignment layer across the 7-layer U-ViT backbone~\cite{uvit2} 
(3 downsampling, 1 bottleneck, and 3 upsampling layers) on RealEstate10K, finding that aligning with earlier blocks yields superior performance.

\noindent{\bf Robustness to Supervision Bias.} To preclude potential ``self-fulfilling'' 
effects, where the model might merely distill the same teacher used for label
generation, we conduct a cross-source robustness study. Specifically, we train 
the baseline using depth labels from Video-Depth-Anything~\cite{VideoDepthAnything}, while our M$^2$-REPA 
is guided by DepthAnythingV2~\cite{depthanytingv2} and DINOv2~\cite{dinov2}. As shown in Tab.~\ref{tab:robustness2}, 
our method consistently yields substantial gains over the baseline on the RealEstate10K 
dataset despite this source mismatch. This confirms that M$^2$-REPA's efficacy transcends simple 
label-space distillation; rather, it enhances generation by regularizing internal 
latent-level representations with structural priors that are orthogonal to 
the specific choice of pseudo-labels.

\begin{table}[H]
\centering
\caption{\textbf{Cross-source robustness analysis on RealEstate10K.} M$^2$-REPA consistently outperforms the baseline under mismatched depth supervision.}
\vspace{-3mm}
{\tiny
\setlength{\tabcolsep}{0.5pt} 
\renewcommand{\arraystretch}{1}
\begin{tabular*}{\textwidth}{@{\extracolsep{\fill}} lc|cccc|cc|c @{}}
\toprule
\multirow{2}{*}{\textbf{Method}} & \multirow{2}{*}{\textbf{Frames}} & \multicolumn{4}{c|}{\textbf{RGB modality}} & \multicolumn{2}{c|}{\textbf{Depth modality}} & \textbf{Mask modality} \\
 & & \textbf{FVD↓} & \textbf{LPIPS↓} & \textbf{SSIM↑} & \textbf{PSNR↑} & \textbf{AbsRel↓} & \textbf{$\boldsymbol{\delta_1}$↑} & \textbf{mIoU↑} \\
\midrule
Video-Depth-Anything Baseline           & 8   & \underline{124.00} & \underline{0.3120} & \underline{0.6690} & \textbf{19.1000} & \underline{0.4530} & \underline{0.5300} & \textbf{0.8810} \\
M$^2$-REPA (DINOv2+DepthAnythingV2)            & 8   & \textbf{118.00} & \textbf{0.2950} & \textbf{0.6800} & \underline{18.7000} & \textbf{0.4170} & \textbf{0.5720} & \underline{0.8800} \\
\midrule
Video-Depth-Anything Baseline           & 200 & \underline{346.75} & \underline{0.5332} & \underline{0.3916} & \underline{12.0521} & \underline{0.5184} & \underline{0.3754} & \underline{0.6109} \\
M$^2$-REPA (DINOv2+DepthAnythingV2)            & 200 & \textbf{291.00} & \textbf{0.5046} & \textbf{0.4223} & \textbf{12.3226} & \textbf{0.5099} & \textbf{0.3989} & \textbf{0.6365} \\
\bottomrule
\end{tabular*}
}
\label{tab:robustness2}
\end{table}


\subsection{Limitations and Broader Impact} M$^2$-REPA's robustness is tied to the quality of the employed foundation experts, which may contain implicit biases. Furthermore, multi-expert alignment requires additional training compute, albeit without inference overhead. Broadly, the dual-use nature of high-fidelity synthesis highlights the importance of developing robust media authentication tools.

\section{Conclusion} 
We presented M$^2$-REPA, the first representation alignment framework for multi-modal video generation that simultaneously integrates complementary priors from diverse foundation experts. By explicitly decoupling modality-specific features and enforcing joint alignment, our method resolves inherent representation conflicts while making use of heterogeneous pre-trained knowledge. Empirical results demonstrate M$^2$-REPA's consistent improvements over competitive baselines across all modalities, providing a strong baseline and a promising direction for multi-modal world modeling.

\section*{Acknowledgements}
This work was partly supported by the Special Foundations for the Development of Strategic Emerging Industries of Shenzhen (No.~KJZD20231023094700001) and the Shenzhen-Tsinghua Special Project for Fundamental \& Frontier Research in Artificial Intelligence (No.~AI2026018).

\clearpage
\appendix
\section*{Supplementary Material}

\makeatletter
\@addtoreset{figure}{section}
\@addtoreset{table}{section}
\@addtoreset{equation}{section}
\makeatother
\setcounter{figure}{0}
\setcounter{table}{0}
\setcounter{equation}{0}
\renewcommand{\thefigure}{\thesection.\arabic{figure}}
\renewcommand{\thetable}{\thesection.\arabic{table}}
\renewcommand{\theequation}{\thesection.\arabic{equation}}

\section{Latent-Space DiT Architecture Details}
\label{sec:supp_dit_arch}

\begin{figure*}[t]
    \centering
    \includegraphics[width=1.00\linewidth]{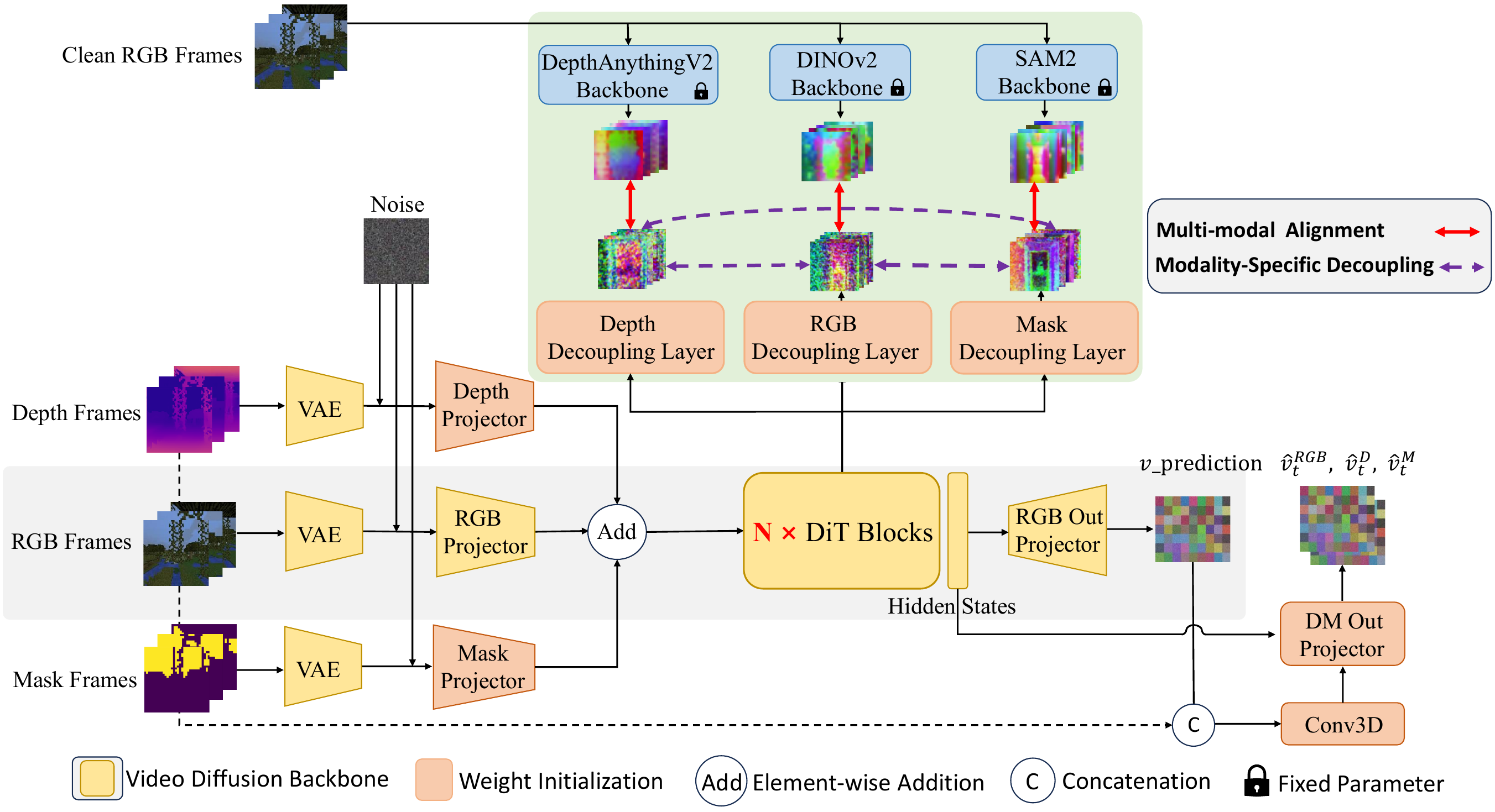}
    \caption{\textbf{Illustration of M$^2$-REPA based on the DiT backbone.} We retain the original RGB projector and design new output projectors for depth and mask.}
    \label{fig:supp_dit_arch}
\end{figure*}

To broaden the evaluation scope of our method and validate its effectiveness in highly dynamic and diverse scenarios, we conduct additional experiments in the Minecraft~\cite{minecraft} environment utilizing a latent-space Diffusion Transformer (DiT)~\cite{dit} as the backbone. As illustrated in \cref{fig:supp_dit_arch}, we build our baseline upon the DFoT architecture~\cite{doft} and adapt it for tri-modal generation.

Unlike the pixel-space instantiation, we first offline encode the Minecraft dataset~\cite{minecraft} across the three modalities into latent representations using a pre-trained VAE~\cite{vae}, yielding $z^m = \text{VAE}(\mathit{X}^m)$ for each modality $m \in \{\text{RGB}, \text{D}, \text{M}\}$. During training, these latents are perturbed by noise to obtain $z_t^m$. For the input design, we duplicate the pre-trained RGB patch projection layer to serve as the dedicated embedders for the newly introduced Depth and Mask modalities. The individual embeddings are then aggregated via element-wise summation to form a unified input token sequence $\mathit{e}_t = \text{Proj}_{\text{in}}^{\text{RGB}}(z_t^{\text{RGB}}) + \text{Proj}_{\text{in}}^{\text{D}}(z_t^{\text{D}}) + \text{Proj}_{\text{in}}^{\text{M}}(z_t^{\text{M}})$. The DiT backbone processes this joint representation, conditioned on the timestep $t$ and action signals $\mathit{a}$, to extract the sequence of hidden states $\mathit{h}_t = f_\theta(\mathit{e}_t, \mathit{a}, t)$. At the output stage, inspired by TesserAct~\cite{tesseract}, we retain the original head to predict the RGB velocity field $\hat{\mathit{v}}_t^{\text{RGB}}$. To address the distinct spatial characteristics of depth and mask, we introduce auxiliary branches: noisy input latents are concatenated with denoised RGB features and routed through dedicated 3D convolutions (Conv3D). These extracted local features are then combined with the global DiT hidden states $\mathit{h}_t$ and passed through DM Output Projectors to yield depth ($\hat{\mathit{v}}_t^{\text{D}}$) and mask ($\hat{\mathit{v}}_t^{\text{M}}$) velocity fields.
To strictly preserve pre-trained knowledge and ensure optimization stability, we initialize our model using the pre-trained single-modality DiT weights. Crucially, all newly introduced structural modules---including the input projectors for depth and mask, the Conv3D layers, and the DM Output Projectors---are zero-initialized. This design guarantees that at the onset of training, the tri-modal model's RGB output perfectly matches that of the original single-modality DiT baseline.

\section{Ablation on Projection Head Design}

\subsection{Architectural Choice and Philosophy}
We employ simple multi-layer perceptrons (MLPs) as our projection heads, guided by three core considerations. 
\textbf{First}, in adherence to the parsimony principle established by REPA~\cite{REPA}, we minimize architectural complexity to ensure that performance gains are strictly attributable to alignment regularization itself, rather than the representational capacity of the projector. 
\textbf{Second}, MLPs provide sufficient non-linear expressivity to map diffusion features $h_t$ into the semantic space of pre-trained encoders. Since our alignment operates at the spatially localized, per-patch level, the global sequence-modeling capability of Transformer-based projectors is both redundant and computationally inefficient. 
\textbf{Third}, the compact nature of MLPs facilitates superior training stability and efficiency—a critical property when serving as an auxiliary regularization term. As shown in Tab.~\ref{tab:projector2}, we empirically compare MLPs against more sophisticated projection mechanisms, including attention-based adapters and query-based decoupling, confirming that lightweight MLPs achieve the optimal trade-off between cross-modal alignment fidelity and computational overhead.

\vspace{-1mm}
\begin{table}[h]
\centering
\caption{\textbf{Ablation on projection head design.} Comparison of attention-based 
adapters, query-based decoupling, and lightweight MLPs (Ours) on short-range (8-frame) 
and long-range (200-frame) generation across RGB, Depth, and Mask modalities. 
Lightweight MLPs achieve the best overall performance with minimal computational cost. 
\textbf{Bold} and \underline{underline} denote the best and second-best results.}
\label{tab:projector2}
\vspace{1.5mm}

\tiny
\setlength{\tabcolsep}{4pt}

\begin{tabular}{lc|cccc|cc|c}
\toprule
\multirow{2}{*}{\textbf{Method}} & \multirow{2}{*}{\textbf{Frames}} & \multicolumn{4}{c|}{\textbf{RGB modality}} & \multicolumn{2}{c|}{\textbf{Depth modality}} & \textbf{Mask} \\
 & & \textbf{FVD↓} & \textbf{LPIPS↓} & \textbf{SSIM↑} & \textbf{PSNR↑} & \textbf{AbsRel↓} & \textbf{$\boldsymbol{\delta_1}$↑} & \textbf{mIoU↑} \\
\midrule
Attention-based adapters           & 8   & 176.00 & 0.2520 & 0.6750 & 14.4000 & 0.9312 & 0.4490 & \underline{0.9012} \\
Query-based decoupling             & 8   & \underline{81.60} & \underline{0.2190} & \underline{0.7270} & \underline{20.2000} & \underline{0.8315} & \textbf{0.4640} & 0.9010 \\
Lightweight MLPs (Ours)            & 8   & \textbf{81.13} & \textbf{0.2123} & \textbf{0.7439} & \textbf{20.4845} & \textbf{0.8224} & \underline{0.4449} & \textbf{0.9045} \\
\midrule
Attention-based adapters           & 200 & 312.50 & 0.4695 & 0.4323 & 12.0828 & 0.9834 & \underline{0.1394} & \underline{0.6574} \\
Query-based decoupling             & 200 & \underline{274.00} & \underline{0.4616} & \textbf{0.4402} & \textbf{13.0215} & \textbf{0.9744} & 0.1314 & 0.6492 \\
Lightweight MLPs (Ours)            & 200 & \textbf{262.50} & \textbf{0.4592} & \underline{0.4374} & \underline{12.8918} & \underline{0.9813} & \textbf{0.1408} & \textbf{0.6591} \\
\bottomrule
\end{tabular}
\vspace{-5.5mm}
\end{table}

\subsection{Projection Head Depth}
We ablate the number of MLP layers used in the projection head. Alongside comparisons against attention-based adapters (Tab.~\ref{tab:mlp_depth_ablation}), we find that a lightweight 3-layer MLP achieves strong and stable alignment performance, while increasing network depth or complexity yields no consistent improvement. This validates our design choice of using simple yet effective projection heads, where alignment quality is governed by the regularization objective rather than projector capacity.

\begin{table}[t]
\centering
\vspace{6mm}
\caption{\textbf{Ablation on projection head depth} for long-range (200-frame) generation. 
A lightweight 3-layer MLP achieves strong and stable performance, 
while deeper networks provide inconsistent gains.}
\label{tab:mlp_depth_ablation}
\vspace{2mm}

\tiny
\setlength{\tabcolsep}{5.0pt}

\begin{tabular}{l|cccc|cc|c}
\toprule
\multirow{2}{*}{\textbf{Method}} & \multicolumn{4}{c|}{\textbf{RGB modality}} & \multicolumn{2}{c|}{\textbf{Depth modality}} & \textbf{Mask modality} \\
 & \textbf{FVD$\downarrow$} & \textbf{LPIPS$\downarrow$} & \textbf{SSIM$\uparrow$} & \textbf{PSNR$\uparrow$} & \textbf{AbsRel$\downarrow$} & \textbf{$\boldsymbol{\delta_1}\uparrow$} & \textbf{mIoU$\uparrow$} \\
\midrule
MLPs (2-layer)   
& 355.25 
& 0.4918 
& 0.4220 
& 12.6197 
& 0.9859 
& 0.1355 
& 0.6330 \\
MLPs (3-layer) (Ours) 
& \underline{262.50} 
& \underline{0.4592} 
& \underline{0.4374} 
& \underline{12.8918} 
& 0.9813 
& 0.1408 
& \underline{0.6591} \\
MLPs (6-layer)   
& 280.50 
& \textbf{0.4580} 
& \textbf{0.4477} 
& \textbf{13.1117} 
& \underline{0.9772} 
& \underline{0.1471} 
& \textbf{0.6672} \\
MLPs (9-layer)   
& \textbf{258.75} 
& 0.4723 
& 0.4239 
& 12.7615 
& \textbf{0.9679} 
& \textbf{0.1685} 
& 0.6408 \\
\bottomrule
\end{tabular}
\vspace{-5mm}
\end{table}

\section{Hyperparameter Sensitivity}

To better demonstrate the impact of the decoupling regularization, we conduct an ablation study on the loss weight hyperparameters. Specifically, following the protocol of REPA~\cite{REPA}, we fix $\lambda_{\text{align}}=0.5$ and systematically sweep the decoupling weight $\lambda_{\text{decouple}} \in \{0.05, 0.5, 35, 100, 200\}$. As shown in Tab.~\ref{tab:lambda_ablation}, the generative performance remains consistently stable across a wide range of $\lambda_{\text{decouple}} \in [0.05, 35]$, demonstrating the inherent robustness of our method to this hyperparameter. However, assigning excessively large values to $\lambda_{\text{decouple}}$ introduces optimization instability and causes the training to diverge from the optimal regime. This observation suggests that a moderate regularization strength is sufficient to achieve effective feature decoupling, without compromising the primary generative objective.

\begin{table}[t]
\centering
\vspace{6mm}
\caption{\textbf{Ablation on the decoupling weight $\lambda_{\text{decouple}}$} 
for long-range (200-frame) generation. Performance remains stable for 
$\lambda_{\text{decouple}} \in [0.05, 35]$ but degrades under excessively large values.}
\label{tab:lambda_ablation}
\vspace{2mm}

\tiny
\setlength{\tabcolsep}{6pt}

\begin{tabular}{l|cccc|cc|c}
\toprule
\multirow{2}{*}{\textbf{Method}} & \multicolumn{4}{c|}{\textbf{RGB modality}} & \multicolumn{2}{c|}{\textbf{Depth modality}} & \textbf{Mask modality} \\
 & \textbf{FVD$\downarrow$} & \textbf{LPIPS$\downarrow$} & \textbf{SSIM$\uparrow$} & \textbf{PSNR$\uparrow$} & \textbf{AbsRel$\downarrow$} & \textbf{$\boldsymbol{\delta_1}\uparrow$} & \textbf{mIoU$\uparrow$} \\
\midrule
$\lambda = 200$ 
& 328.75 
& \underline{0.4689} 
& \underline{0.4340} 
& \underline{12.7331} 
& 0.9939 
& 0.1275 
& \underline{0.6470} \\
$\lambda = 100$ 
& 310.00 
& 0.4787 
& 0.4174 
& 12.2055 
& \underline{0.9926} 
& \underline{0.1285} 
& 0.6114 \\
$\lambda = 35$  
& 287.00 
& 0.4733 
& 0.4221 
& 12.4622 
& 0.9958 
& 0.1246 
& 0.6301 \\
$\lambda = 0.5$  
& \underline{284.25} 
& 0.4731 
& 0.4226 
& 12.4616 
& 0.9959 
& 0.1246 
& 0.6284 \\
$\lambda = 0.05$ (Ours) 
& \textbf{262.50} 
& \textbf{0.4592} 
& \textbf{0.4374} 
& \textbf{12.8918} 
& \textbf{0.9813} 
& \textbf{0.1408} 
& \textbf{0.6591} \\
\bottomrule
\end{tabular}
\vspace{-5mm}
\end{table}

\section{Qualitative Comparison in Minecraft Environment}

\begin{figure}[h]
    \centering
    \includegraphics[width=0.96\linewidth]{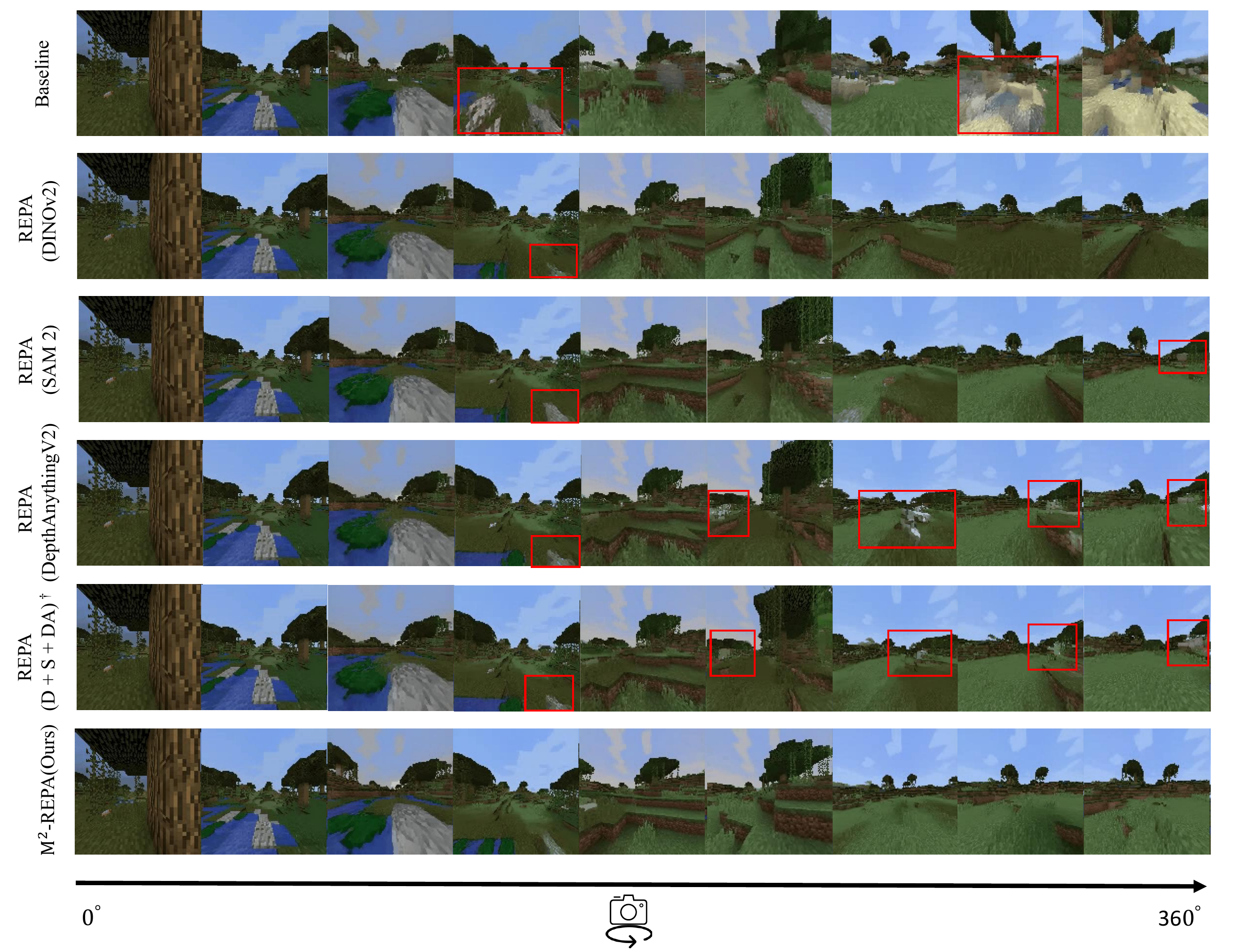}
    \vspace{-0.3\intextsep} 
    \caption{
    Qualitative comparison of long-horizon action-conditioned video generation (150 frames) in the Minecraft environment.
    Each video is generated from an initial frame observation and the corresponding per-frame action sequence.
    Our M$^2$-REPA maintains stable scene coherence and temporal consistency throughout the generation horizon. $^\dagger$D+S+DA denotes the naive combination of DINOv2, SAM2, and DepthAnythingV2.
    }
    \label{fig:rgbdseg-visual22}
    \vspace{-0.7\intextsep} 
\end{figure}

Figure~\ref{fig:rgbdseg-visual22} presents a long-horizon qualitative comparison (150 frames) between our \textbf{M$^2$-REPA} and baseline methods in the Minecraft environment. Each video sequence is generated conditioned on a single first-frame RGB observation along with future action instructions.

As illustrated in Figure~\ref{fig:rgbdseg-visual22}, both the RGB-DM baseline and the unimodal REPA variant suffer from varying degrees of scene collapse as generation progresses. The RGB-DM baseline, lacking explicit multimodal representation alignment, fails to preserve global scene-level semantics over extended temporal horizons, leading to severe structural distortions and accumulated visual artifacts in later frames. Similarly, despite leveraging single-modality representation alignment, the unimodal REPA variant still exhibits noticeable temporal inconsistencies, suggesting that aligning representations from a single modality alone is insufficient for robust long-horizon video generation.
In contrast, M$^2$-REPA consistently produces high-fidelity RGB frames that maintain stable scene coherence and temporal consistency throughout the entire 150-frame horizon. By effectively exploiting complementary semantic cues across multiple modalities, our multimodal representation alignment strategy enables the diffusion model to sustain a coherent internal world representation, yielding temporally stable video sequences even under challenging long-horizon settings. These results further demonstrate the effectiveness of our proposed framework.

\section{Mask Segmentation Evaluation Metric}

We evaluate mask generation quality using mean Intersection over Union (mIoU) with a greedy matching protocol. 
For a frame containing predicted masks $\{P_i\}_{i=1}^{N}$ and ground-truth masks $\{G_j\}_{j=1}^{M}$, where $P_i, G_j \in \{0,1\}^{H \times W}$, we first compute pairwise IoU as $\text{IoU}(P_i, G_j) = |P_i \cap G_j| / |P_i \cup G_j|$.
We then establish one-to-one correspondences through greedy matching: for each ground truth mask $G_j$, we select the unmatched prediction $P_i$ with maximum IoU and mark them as matched if $\text{IoU}(P_i, G_j) > 0.5$. 

The frame-level mIoU is computed as:
\begin{equation}
\text{mIoU}_{\text{frame}} = \frac{1}{K} \sum_{k=1}^{K} \text{IoU}(P_{i_k}, G_{j_k}),
\end{equation}
where $K$ is the number of matched pairs. When $K=0$, we set $\text{mIoU}_{\text{frame}} = 0$. 
The overall mIoU averages frame-level scores across all non-context frames in the dataset:
\begin{equation}
\text{mIoU}_{\text{overall}} = \frac{1}{B \times F} \sum_{b=1}^{B} \sum_{f=1}^{F} \text{mIoU}_{\text{frame}}^{(b,f)}
\end{equation}

Predicted probability maps are binarized at threshold 0.5. 
Context frames used for conditioning are excluded from evaluation. 
Our greedy matching provides comparable accuracy to optimal Hungarian matching while being computationally efficient.

\section{Generalization to Independently-Annotated Ground Truth}
\label{sec:supp_omniworld}

To verify that the gains of M$^2$-REPA are not an artifact of the coupling between the alignment targets and the evaluation references, we retrain and re-evaluate our method on OmniWorld-Game~\cite{omniworld}, a recent multi-modal dataset whose annotations are obtained independently of our alignment teachers. In this setting, the depth maps are engine-rendered ground truth, and the instance masks are produced by full-scene GroundingDINO$+$SAM annotations rather than the per-frame predictions used during training. As reported in Tab.~\ref{tab:omniworld}, M$^2$-REPA continues to outperform the RGB-DM baseline, the single-teacher REPA variant, and the naive multi-teacher combination across the generative (FVD, LPIPS, PSNR), depth (AbsRel, $\delta_1$), and mask (mIoU) metrics. Notably, the advantage persists even though the depth and mask references here are decoupled from the alignment supervision, indicating that the improvements stem from the complementary multi-modal priors internalized during training rather than from any leakage between the alignment targets and the evaluation protocol.

\begin{table}[h]
\centering
\caption{\textbf{Generalization to OmniWorld-Game~\cite{omniworld}}, whose annotations are decoupled from our alignment teachers, under long-range (200-frame) generation. \textbf{Bold} denotes the best result.}
\label{tab:omniworld}
\vspace{1.5mm}

\tiny
\setlength{\tabcolsep}{5pt}

\begin{tabular}{l|ccc|cc|c}
\toprule
\multirow{2}{*}{\textbf{Method}} & \multicolumn{3}{c|}{\textbf{RGB modality}} & \multicolumn{2}{c|}{\textbf{Depth modality}} & \textbf{Mask} \\
 & \textbf{FVD$\downarrow$} & \textbf{LPIPS$\downarrow$} & \textbf{PSNR$\uparrow$} & \textbf{AbsRel$\downarrow$} & \textbf{$\boldsymbol{\delta_1}\uparrow$} & \textbf{mIoU$\uparrow$} \\
\midrule
RGB-DM Baseline                       & 217.0 & 0.355 & 18.48 & 7.141 & 0.124 & 0.541 \\
REPA (DepthAnythingV2)                & 175.5 & 0.326 & 19.30 & 2.429 & 0.124 & 0.607 \\
REPA (D+S+DA, naive)                  & 198.6 & 0.345 & 18.81 & 2.984 & \textbf{0.127} & 0.549 \\
\textbf{M$^2$-REPA (Ours)}            & \textbf{157.4} & \textbf{0.318} & \textbf{19.96} & \textbf{0.979} & 0.124 & \textbf{0.616} \\
\bottomrule
\end{tabular}
\end{table}

\section{Scaling to Large-Scale Models}
\label{sec:supp_scale}

M$^2$-REPA is a plug-and-play alignment recipe that is agnostic to the underlying generator. To verify its applicability to large-scale models, we apply M$^2$-REPA to TesserAct~\cite{tesseract}, a 4D world model built upon the CogVideoX-5B~\cite{cogvideox} backbone, and evaluate on RealEstate10K. As reported in Tab.~\ref{tab:tesseract}, M$^2$-REPA delivers the best results across all metrics, lowering FVD from $639.3$ for the TesserAct baseline to $418.3$ while simultaneously improving perceptual quality (LPIPS, PSNR) and depth accuracy (AbsRel, $\delta_1$).

\begin{table}[h]
\centering
\caption{\textbf{Scaling M$^2$-REPA to TesserAct (CogVideoX-5B)~\cite{tesseract} on RealEstate10K.} M$^2$-REPA remains the best across all metrics on a 5B-parameter backbone. \textbf{Bold} and \underline{underline} denote the best and second-best results.}
\label{tab:tesseract}
\vspace{1.5mm}

\tiny
\setlength{\tabcolsep}{6pt}

\begin{tabular}{l|ccc|cc}
\toprule
\multirow{2}{*}{\textbf{Method}} & \multicolumn{3}{c|}{\textbf{RGB modality}} & \multicolumn{2}{c}{\textbf{Depth modality}} \\
 & \textbf{FVD$\downarrow$} & \textbf{LPIPS$\downarrow$} & \textbf{PSNR$\uparrow$} & \textbf{AbsRel$\downarrow$} & \textbf{$\boldsymbol{\delta_1}\uparrow$} \\
\midrule
TesserAct Baseline~\cite{tesseract}     & 639.3 & 0.6255 & 11.20 & 0.591 & 0.237 \\
REPA (DINOv2)                           & \underline{475.0} & 0.5243 & \underline{13.47} & 0.591 & 0.250 \\
REPA (Multi-expert naive)               & 564.8 & \underline{0.5242} & 12.17 & \underline{0.569} & \underline{0.257} \\
\textbf{M$^2$-REPA (Ours)}              & \textbf{418.3} & \textbf{0.4417} & \textbf{14.68} & \textbf{0.565} & \textbf{0.270} \\
\bottomrule
\end{tabular}
\end{table}

\section{Centered Kernel Alignment (CKA)}
\label{sec:supp_cka}

We provide a mathematical overview of Centered Kernel Alignment (CKA)~\cite{CKA}, the similarity metric used in our modality-specific decoupling regularization loss $\mathcal{L}_{\text{decouple}}$ (Eq. 7 in the main paper).

CKA is a representation similarity measure based on the Hilbert-Schmidt Independence Criterion (HSIC)~\cite{CKA}. Given two feature matrices $X \in \mathbb{R}^{n \times p_1}$ and $Y \in \mathbb{R}^{n \times p_2}$ with $n$ samples, linear CKA is computed by first constructing Gram matrices $K = XX^T$ and $L = YY^T$, then centering them using $H = I - \frac{1}{n}\mathbf{1}\mathbf{1}^T$ to obtain $\tilde{K} = HKH$ and $\tilde{L} = HLH$. The normalized similarity is then:

\begin{equation}
\begin{aligned}
\text{CKA}(X, Y) &= \frac{\text{HSIC}(K, L)}{\sqrt{\text{HSIC}(K, K) \cdot \text{HSIC}(L, L)}} \\
&= \frac{\|Y^TX\|_F^2}{\|X^TX\|_F \cdot \|Y^TY\|_F},
\end{aligned}
\label{eq:cka_definition}
\end{equation}
where $\text{HSIC}(K, L) = \frac{1}{(n-1)^2}\text{tr}(\tilde{K}\tilde{L})$ and $\|\cdot\|_F$ denotes the Frobenius norm.

CKA is well-suited for multi-modal feature decoupling due to its key properties: (1) bounded range $[0, 1]$ facilitating optimization, (2) invariance to orthogonal transformations ($\text{CKA}(XQ, Y) = \text{CKA}(X, Y)$ for $Q^TQ = I$), and (3) invariance to isotropic scaling ($\text{CKA}(aX, bY) = \text{CKA}(X, Y)$). These properties enable robust comparison of features from heterogeneous foundation models regardless of arbitrary rotations or magnitude differences. By minimizing pairwise CKA among decoupled features $\{\hat{h}^{(k)}_\phi\}_{k=1}^K$, we explicitly encourage orthogonality and complementarity across modalities, preventing feature conflicts during multi-modal alignment.

\bibliographystyle{splncs04}
\bibliography{main}

\end{document}